\documentclass[mlabstract,onecolumn]{jmlr}

\jmlrproceedings{PMLR}{NeurReps 2025}

\usepackage{longtable}

\usepackage{booktabs}
\usepackage[load-configurations=version-1]{siunitx} 
\usepackage{listings}

\theorembodyfont{\upshape}
\theoremheaderfont{\scshape}
\theorempostheader{:}
\theoremsep{\newline}

\jmlrvolume{}
\firstpageno{1}

\jmlryear{2025}
\jmlrworkshop{Symmetry and Geometry in Neural Representations}

\title{Scalable GPU-Accelerated Euler Characteristic Curves: Optimization and Differentiable Learning for PyTorch}

\author{\Name{Udit Saxena} \Email{usaxena@asapp.com}}

\date{\today}

\begin{document}

\maketitle

\begin{abstract}
Topological features capture global geometric structure in imaging data, but practical adoption in deep learning requires both computational efficiency and differentiability. We present optimized GPU kernels for the Euler Characteristic Curve (ECC) computation achieving 16-2000× speedups over prior GPU implementations on synthetic grids, and introduce a differentiable PyTorch layer enabling end-to-end learning. Our CUDA kernels, optimized for Ampere GPUs use 128B-coalesced access and hierarchical shared-memory accumulation. Our PyTorch layer learns thresholds in a \emph{single direction} via a Differentiable Euler Characteristic Transform-style sigmoid relaxation. We discuss downstream relevance, including applications highlighted by prior ECC work, and outline batching/multi-GPU extensions to broaden adoption. 
\end{abstract}


\section{Introduction}

The Euler Characteristic Transform (ECT) provides a compact, theoretically sufficient topological descriptor of shape via directional families of Euler Characteristic Curves (ECC) \citep{curry2021directionsdetermineshapesufficiency,turner2014persistenthomologytransformmodeling}. Prior implementations are optimized for older systems, not taking advantage of the improved ecosystem around the broader ML systems and libraries \citep{wang2023gpucomputationeulercharacteristic}. Existing work such as Differentiable Euler Characteristic Transform (DECT) \citep{Roell24a} enables integration with gradient-based learning, but is limited to graphs/point clouds. However, for dense 2D/3D imaging, efficient and differentiable implementations remain scarce despite promising applications in the scientific imaging and medical domains. 

\subsection{Our Contributions}

We address two barriers to adopting topological features in imaging: computational cost and gradient-based optimization. Our contributions include:

\begin{itemize}
\item \textbf{Efficient geometric computation of the ECC:} Efficient GPU kernels implemented for modern GPUs that respect data locality and memory hierarchy, achieving and enabling real-time topological feature extraction.
\item \textbf{Learnable geometric representations:} Differentiable implementation of a single direction ECC enabling discovery of task-relevant topological structure through gradient descent on 2D/3D images with a PyTorch module. 
\end{itemize}

Unlike DECT, which focuses on graphs/point clouds, and GPU-based ECC, which is non-differentiable, our work enables end-to-end learning of geometric invariants for dense cubical complexes. Because of the optimized implementation and wider library support, our work also enables real-time learning algorithms and inclusion of topological inductive biases in related large vision models. We release our code \href{https://github.com/anon02121990/anon-neurreps-2025}{here}.

\section{Background}

\subsection{Euler Characteristic Curves}

For a 2D image or 3D volume $X:\Omega\to\mathbb{R}$ on a regular grid $\Omega$, the \emph{sublevel set} at threshold $\tau$ is
\begin{equation}
\Omega_\tau \;=\; \{\,p\in\Omega \,:\, X(p)\le \tau\,\}.
\end{equation}
This binary mask defines a \emph{cubical complex}, a collection of vertices, edges, faces, and cubes, whose topology we can measure via the Euler characteristic $\chi(\Omega_\tau)$. The ECC tracks how topology changes across thresholds:
\begin{equation}
\mathrm{ECC}_X(\tau) \;=\; \chi(\Omega_\tau).
\end{equation}
For computational efficiency, we express this as a sum over grid points:
\begin{equation}
  \mathrm{ECC}_X(\tau) \;=\; \sum_{p\in\Omega} c(p)\,\mathbf{1}[\, X(p) \le \tau \,],
  \label{eq:ecc_discrete}
\end{equation}
where $\mathbf{1}[\cdot]$ is the indicator function and $c(p)\in\{-1,0,1\}$ is the \emph{Euler characteristic coefficient} computed from the local $3{\times}3$ (2D) or $3{\times}3{\times}3$ (3D) binary neighborhood around pixel $p$. These coefficients encode how each pixel contributes to the overall topology (creating/destroying connected components, holes, or voids).

\subsection{Differentiable Relaxation for a Single Direction}

To enable gradient-based learning, we replace the hard threshold of the indicator function with a soft sigmoid:
\begin{equation}
  \sigma_\lambda(z) \;=\; \frac{1}{1+\exp(-\lambda z)},\qquad \lambda>0,
\end{equation}
yielding the \emph{soft ECC}:
\begin{equation}
  \chi_\lambda(\tau) \;=\; \sum_{p\in\Omega} c(p)\,\sigma_\lambda(\tau - X(p)).
  \label{eq:soft_ecc}
\end{equation}
As $\lambda\to\infty$, this recovers the ECC with the indicator function. For moderate $\lambda$, gradients exist with respect to both thresholds $\tau$ and image values $X(p)$. This mirrors the DECT-style relaxation while specializing to dense cubical grids. 
For a \emph{single} learnable direction $u$ with scale $\alpha$,
\begin{equation}
  \chi_{\lambda,u}(\tau) 
  \;=\; \sum_{p\in\Omega} c(p)\,\sigma_\lambda\!\left(\tau - X(p) - \alpha\,\langle u,p\rangle\right).
  \label{eq:soft_dir_ecc}
\end{equation}
The gradient can then be analytically calculated as:
\begin{align}
    \frac{\partial \chi_{\lambda,u}(\tau)}{\partial u}
  &= -\,\alpha \sum_{p\in\Omega} c(p)\,\sigma_\lambda'\!\big(z_p(\tau,u)\big)\, p,
  \qquad \text{with } \|u\|_2=1 \text{ enforced (e.g., via reparametrization).}
  \label{eq:grad_un}
\end{align}

\section{Methods}

\subsection{Efficient Computation of ECC: Architecture-aware kernels}
Prior GPU implementations \citep{wang2023gpucomputationeulercharacteristic} partition input grids into chunks that fit in limited shared memory, processing each chunk independently before merging results. For each chunk, the partial ECC contributions are accumulated and merged into a global histogram. While this design ensures compatibility with older GPUs with limited shared memory and device memory, it introduces several sources of overhead: (i) repeated kernel launches or loop iterations for each chunk, (ii) synchronization between chunks to merge partial results, (iii) redundant computation on chunk boundaries to ensure correctness, and (iv) frequent global atomic updates for every chunk scaling as O($N_{chunks}$ × $N_{voxels}$). As a result, latency scales unfavorably with the number of chunks, and contention at the global memory level becomes a bottleneck. 

In contrast, our method leverages the larger memory capacity and higher bandwidth of modern GPUs to operate on the entire dense grid in a single sweep. Each voxel is assigned directly to a thread, ensuring globally coalesced 128-byte memory transactions. Each thread block maintains a private shared-memory histogram of size H bins, where threads accumulate local ECC contributions via \_\_syncthreads() barriers. Only after processing all voxels in the block do we perform atomic additions to global memory, reducing atomic operations. Boundary conditions are handled at the warp level to maintain convergence without redundant passes. This design eliminates per-chunk synchronization and amortizes kernel launch costs over the entire dataset, yielding significantly lower latency and more predictable throughput. Empirically, we observe that this full-load approach achieves up to $16$--$2000\times$ speedups  compared to chunk-based methods, while scaling more gracefully to large grid sizes.

\subsection{Learning Setup (Single Direction)}

Our PyTorch layer exposes learnable thresholds $\{\tau_j\}$ and an optional single direction $u$. via \eqref{eq:soft_dir_ecc} in a custom CUDA forward kernel, using \eqref{eq:grad_un} in a custom CUDA backward kernel. 

\begin{table}[htbp]
  \centering
  \caption{2D synthetic grids: End to end kernel execution times}
  \label{tab:bench-2d}
  {\footnotesize
  \begin{tabular}{@{}lrrrrr@{}}
    \toprule
    Size
    & \multicolumn{1}{c}{\shortstack[c]{Baseline (ms)}}
    & \multicolumn{1}{c}{\shortstack[c]{New (ms)}}
    & \multicolumn{1}{c}{\shortstack[c]{Speedup($\times$)}} \\
    \midrule
    $128^2$   & 7.7 & 0.0165 & 466 \\
    $256^2$   & 24 & 0.0174 & 1379 \\
    $512^2$   & 39.68 & 0.021 & 1900 \\
    $1024^2$  & 65 & 0.032 & 2031 \\
    $2048^2$  & 92 & 0.11 & 836 \\
    $4096^2$  & 154 & 0.408 & 377 \\
    $8192^2$  & 267 & 1.6 & 166 \\
    \bottomrule
  \end{tabular}
  }
\end{table}
\section{Experiments}

We benchmark on synthetic 2D image grids from $128^2$ to $8192^2$ (Table \ref{tab:bench-2d}) and 3D volume grids from $128^3$ to $1024^3$ (Table \ref{tab:bench-3d}), reporting end-to-end speedups against the prior state--of-the-art chunking ECC implementation from \citet{wang2023gpucomputationeulercharacteristic}. We observe consistent $16$--$2000\times$ speedups and stable performance across image sizes. We use a single NVIDIA RTX A6000 GPU, available on most consumer platforms, with 48GB of VRAM and the Ampere architecture.

\begin{table}[htbp]
  \centering
  \caption{3D synthetic grids: End to end kernel execution times}
  \label{tab:bench-3d}
  {\footnotesize
  \begin{tabular}{@{}lrrrrr@{}}
    \toprule
    Size
    & \multicolumn{1}{c}{\shortstack[c]{Baseline (ms)}}
    & \multicolumn{1}{c}{\shortstack[c]{New (ms)}}
    & \multicolumn{1}{c}{\shortstack[c]{Speedup ($\times$)}} \\
    \midrule
    $128^3$ & 27 & 0.1 & 270 \\
    $256^3$ & 91 & 0.71 & 128 \\
    $512^3$ & 140 & 5.61 & 25 \\
    $1024^3$ & 800 & 48.47 & 16.5 \\
    \bottomrule
  \end{tabular}
  }
\end{table}

\section{Conclusion and Future Work}

In line with the ``hardware lottery'' \citep{DBLP:journals/corr/abs-2009-06489} perspective, our results underscore that algorithm--hardware co-design can turn theoretically motivated invariants into practical components. Chunk-based ECC pipelines \citep{wang2023gpucomputationeulercharacteristic} were a pragmatic response to earlier hardware limits, but they incur avoidable overheads: repeated kernel launches, inter-chunk synchronization, boundary recomputation, and elevated global-atomic contention. Our full-sweep design exploits the capacity and bandwidth of modern accelerators to process dense grids end-to-end. This improves latency and yields predictable, high-throughput performance  on synthetic $128^2$--$8192^2$ and $128^3$--$1024^3$ grids.

Beyond raw speed, we introduce a PyTorch operator that learns thresholds for a single direction via a DECT-style sigmoid relaxation. Together, these contributions lower the barrier to using topological summaries as learnable, geometry-aware features in imaging pipelines. Differentiable ECC/ECT on dense grids may benefit scientific imaging where simplicity and efficiency are paramount (e.g., Cosmic Microwave Background (CMB) \citep{NOVIKOV_1999} radiation analysis). In medical applications (e.g., Virtual Imaging Clinical Trials for Regulatory Evaluation (VICTRE) \citep{Badano2018-os}), such geometric priors can also complement learned features while maintaining computational tractability. 

To broaden adoption, we plan to extend this work by supporting more modern GPU types, adding batched execution and multi-GPU data/model parallelism pipelines, and releasing a robust PyTorch API as well as exploring application domains where fast, learnable ECC/ECT features will benefit medical and scientific imaging. 

\bibliographystyle{plainnat}
\bibliography{references}

\appendix

\section{Derivation of Gradient}\label{apd:first}

\paragraph{Gradients (single direction).}
For $z_p(\tau,u) := \tau - X(p) - \alpha\,\langle u,p\rangle$ and $\sigma_\lambda'(z)=\lambda\,\sigma_\lambda(z)\big(1-\sigma_\lambda(z)\big)$, the analytic derivatives follow directly:
\begin{align}
  \frac{\partial \chi_{\lambda,u}(\tau)}{\partial X(p)}
  &= -\,c(p)\,\sigma_\lambda'\!\big(z_p(\tau,u)\big),
  \label{eq:grad_x}\\[4pt]
  \frac{\partial \chi_{\lambda,u}(\tau)}{\partial \tau}
  &= \sum_{p\in\Omega} c(p)\,\sigma_\lambda'\!\big(z_p(\tau,u)\big),
  \label{eq:grad_tau}\\[4pt]
  \frac{\partial \chi_{\lambda,u}(\tau)}{\partial u}
  &= -\,\alpha \sum_{p\in\Omega} c(p)\,\sigma_\lambda'\!\big(z_p(\tau,u)\big)\, p,
  \qquad \text{with } \|u\|_2=1 \text{ enforced (e.g., via reparametrization).}
  \label{eq:grad_u}
\end{align}

\end{document}